\documentclass[10pt,twocolumn,letterpaper]{article}

\usepackage{cvpr}
\usepackage{times}
\usepackage{epsfig}
\usepackage{graphicx}
\usepackage{amsmath}
\usepackage{amssymb}
\usepackage{booktabs}

\newcommand*{\affmark}[1][*]{\textsuperscript{#1}}
\newcommand*{\affaddr}[1]{#1} 

\usepackage[pagebackref=true,breaklinks=true,letterpaper=true,colorlinks,bookmarks=false]{hyperref}

\cvprfinalcopy 

\ifcvprfinal\fi
\begin{document}

\title{Weakly Supervised Complementary Parts Models for Fine-Grained Image Classification from the Bottom Up}



\author{
Weifeng Ge\affmark[1,2]\thanks{These authors have equal contribution.} \hspace{0.6in} Xiangru Lin\affmark[1]\footnotemark[1] \hspace{0.6in} Yizhou Yu\affmark[2] \\
\affaddr{\affmark[1]The University of Hong Kong}
\hspace{0.6in}
\affaddr{\affmark[2]Deepwise AI Lab}\\
}

\maketitle

\begin{abstract}
Given a training dataset composed of images and corresponding category labels, deep convolutional neural networks show a strong ability in mining discriminative parts for image classification. However, deep convolutional neural networks trained with image level labels only tend to focus on the most discriminative parts while missing other object parts, which could provide complementary information. In this paper, we approach this problem from a different perspective. We build complementary parts models in a weakly supervised manner to retrieve information suppressed by dominant object parts detected by convolutional neural networks. Given image level labels only, we first extract rough object instances by performing weakly supervised object detection and instance segmentation using Mask R-CNN and CRF-based segmentation. Then we estimate and search for the best parts model for each object instance under the principle of preserving as much diversity as possible. In the last stage, we build a bi-directional long short-term memory (LSTM) network to fuze and encode the partial information of these complementary parts into a comprehensive feature for image classification.
Experimental results indicate that the proposed method not only achieves significant improvement over our baseline models, but also outperforms state-of-the-art algorithms by a large margin (6.7\%, 2.8\%, 5.2\% respectively) on Stanford Dogs 120, Caltech-UCSD Birds 2011-200 and Caltech 256.
\end{abstract}

\section{Introduction}
Deep neural networks have demonstrated its ability to learn representative features for image classification~\cite{russakovsky2015imagenet, krizhevsky2012imagenet, simonyan2014very, szegedy2015going, he2016deep}. Given training data, image classification~\cite{everingham2010pascal, krizhevsky2012imagenet} often builds a feature extractor that accepts an input image and a subsequent classifier that generates prediction probability for the image. This is a common pipeline in many high-level vision tasks, such as object detection~\cite{felzenszwalb2008discriminatively, girshick2014rich,he2017mask}, tracking~\cite{Teng_2017_ICCV,Ristani_2018_CVPR,Sun_2018_CVPR}, and scene understanding~\cite{Dvornik_2017_ICCV,Lu_2018_CVPR}.

\begin{figure}[ht]
  \centering
  \includegraphics[width=1.0\linewidth]{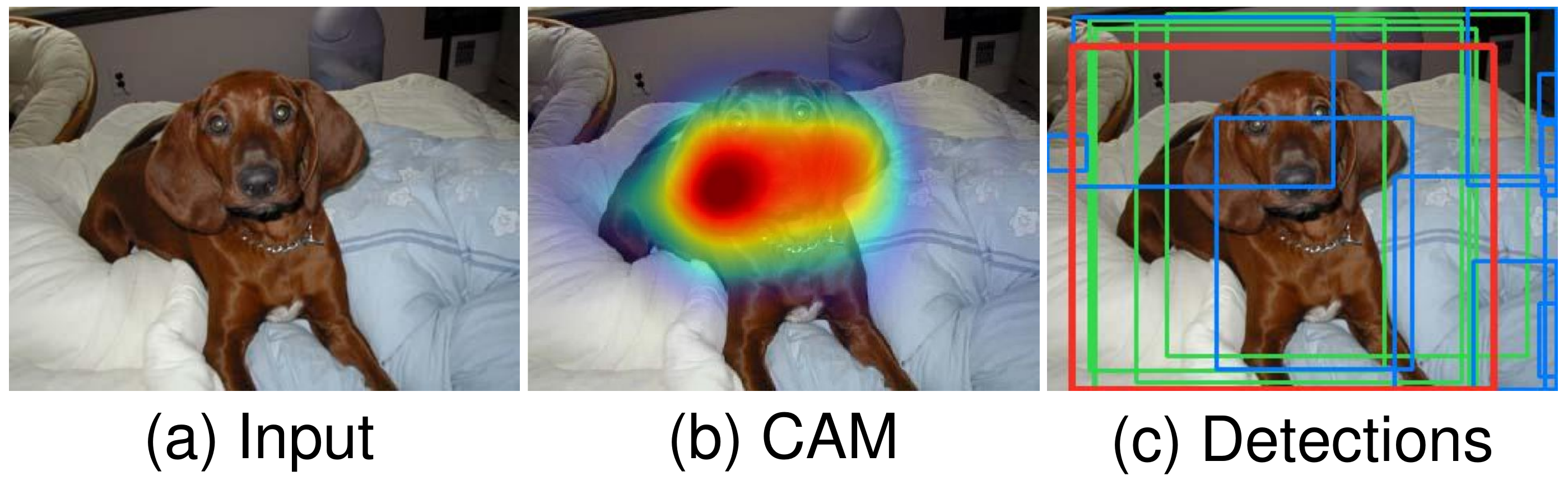}
  \caption{Visualization of class activation map (CAM~\cite{zhou2016learning}) and weakly supervised object detections.}
  \label{Fig:Visualization}
\end{figure}

Although a model trained with the aforementioned pipeline can achieve competitive results on many image classification benchmarks, its performance gain primarily comes from the model's capacity to discover the most discriminative parts in the input image. To better understand a trained deep neural network and obtain insights about this phenomenon, many techniques~\cite{bach2015pixel,zhou2016learning,bau2017network} have been proposed to visualize the intermediate results of deep networks. In Fig \ref{Fig:Visualization}, it can be found that deep convolutional neural networks trained with image labels only tend to focus on the most discriminative parts while missing other object parts. However, focusing on the most discriminative parts alone can have limitations. Some image classification tasks need to grasp object descriptions that are as complete as possible. A complete object description does not have to come in one piece, but could be assembled together using multiple partial descriptions. To remove redundancies, such partial descriptions should be complementary to each other. Image classification tasks, that could benefit from such complete descriptions, include fine-grained classification tasks on Stanford Dogs 120~\cite{KhoslaYaoJayadevaprakashFeiFei_FGVC2011} and CUB 2011-200~\cite{WelinderEtal2010}, where appearances of different object parts collectively contribute to the final classification performance.

According to the above analysis, we approach image classification from a different perspective and propose a new pipeline that aims to mine complementary parts instead of the aforementioned most discriminative parts, and fuse the mined complementary parts before making final classification decisions.

\noindent\textbf{Object Detection Phase.} Object detection~\cite{felzenszwalb2008discriminatively, girshick2014rich,he2017mask} is able to localize objects by performing a huge number of classifications at a large number of locations. In Fig \ref{Fig:Visualization}, the red bounding boxes are the ground truth, the green ones are positive object proposals, and the blue ones are negative proposals. The differences between the positive and negative proposals are whether they contain sufficient information (overlap ratio with the ground truth bounding box) to describe objects. If we look at the activation map in Fig \ref{Fig:Visualization}, it is obvious that the positive bounding boxes spread much wider than the core regions. As a result, we hypothesize that { \em the positive object proposals that lay around the core regions can be helpful for image classification since they contain partial information of the objects in the image}.
However, the challenges in improving image classification by detection are two-fold. First, how can we perform object detection without groundtruth bounding box annotations? Second, how can we exploit object detection results to boost the performance of image classification? In this paper, we attempt to tackle these two challenges in a weakly supervised manner.

To avoid missing any important object parts, we propose a weakly supervised object detection pipeline regularized by iterative object instance segmentation. We start by training a deep classification neural network that produces a class activation map (CAM) as in~\cite{zhou2016learning}. Then the activations in CAM are taken as the pixelwise probabilities of the corresponding class. A conditional random field (CRF)~\cite{sutton2012introduction} then incorporates low level pairwise appearance information to perform unsupervised object instance segmentation. To refine object locations and pixel labels, a Mask R-CNN~\cite{he2017mask} is trained using the object instance masks from the CRF. Results from the Mask R-CNN are used as a pixel probability map to replace the CAM in the CRF. We alternate Mask R-CNN and CRF regularization a few times to generate the final object instance masks.

\noindent\textbf{Image Classification Phase.} Directly reporting classification results in the object detection phase gives rise to inferior performance because object detection algorithms make much effort to determine location in addition to class labels. In order to mine representative object parts with the help of object detection, we utilize the proposals generated in the previous object detection phase and build a complementary parts model, which consists of a subset of the proposals that cover as much complementary object information as possible. At the end, we exploit a bi-directional long short-term memory network to encode the deep features of the object parts for final image classification.

The proposed weakly supervised complementary parts model has been efficiently implemented in Caffe~\cite{jia2014caffe}. Experimental results demonstrate state-of-the-art performance on multiple image classification tasks, including fine-grained classification on Stanford Dogs 120~\cite{KhoslaYaoJayadevaprakashFeiFei_FGVC2011} and Caltech-UCSD Birds 200-2011~\cite{WelinderEtal2010}, and generic classification on Caltech 256~\cite{griffin2007caltech}.

In summary, this paper has the following contributions:
{\flushleft $\bullet$} We introduce a new representation for image classification, called weakly supervised complementary parts model, that attempts to grasp complete object descriptions using a selected subset of object proposals. It is an important step forward in exploiting weakly supervised detection to boost image classification performance.

{\flushleft $\bullet$} We develop a novel pipeline for weakly supervised object detection and instance segmentation. Specifically, we iterate the following two steps, object detection and segmentation using Mask R-CNN, and instance segmentation enhancement using CRF.

{\flushleft $\bullet$} To encode complementary information in different object parts, we exploit a bi-directional long short-term memory network to make the final classification decision. Experimental results demonstrate that we achieve state-of-the-art performance on multiple image classification tasks.
\section{Related Work}
\noindent\textbf{Weakly Supervised Object Detection and Segmentation.} Weakly supervised object detection and segmentation respectively locates and segments objects with image label only~\cite{diba2016weakly}. 
In \cite{durand2016weldon,durand2017wildcat}, the object detection is solved as a classification problem by specific pooling layers in CNNs.  The method in \cite{Wang_2018_CVPR} proposed an iterative bottom-up and top-down framework to expand object regions and optimize segmentation network iteratively. Ge {\em et al.} in ~\cite{Ge_2018_CVPR} progressively mine the object locations and pixel labels with the filtering and fusion of multiple evidences.

While here we perform the weakly supervised object instance detection and segmentation by feeding a coarse segmentation mask and proposal for Mask R-CNN~\cite{he2017mask} using CAM~\cite{zhou2016learning} and rectifying the object locations and masks with CRF~\cite{sutton2012introduction} iteratively. In this way, we avoid losing important object parts for subsequent object parts modeling.


\noindent\textbf{Part Based Fine-grained Image Classification.} Learning a diverse collection of discriminative parts in a supervised\cite{zhang2013deformable,zhang2014part} or unsupervised manner~\cite{simon2015neural,zheng2017learning,lam2017fine} is very popular in fine-grained image classification. Many works~\cite{zhang2013deformable,zhang2014part} have been done to build object part models with part bounding box annotations. The method in ~\cite{zhang2013deformable} builds two deformable part models~\cite{felzenszwalb2008discriminatively} to localize objects and discriminative parts. Zhang {\em et al.} in ~\cite{zhang2014part} treats objects and semantic parts equally by assigning them in different object classes with R-CNN~\cite{girshick2014rich}. Another line of works~\cite{simon2015neural,zheng2017learning,lam2017fine,Wang_2018_CVPR} estimate the part location in a unsupervised setting. In ~\cite{simon2015neural}, parts are discovered based the neural activation, and then are optimized using a EM similar algorithm. The work in~\cite{simon2015neural} extracts the highlight responses in CNN as the part prior to initialize convolutional filters, and then learn discriminative patch detectors end-to-end.

In this paper, we do not aim to build strong part detectors to provide local appearance information for the final classification decision. The goal of our complementary parts model is to efficiently utilize the rich information hidden in the object proposals produced during object detection phase. Every object proposal contains enough information to classify the object, and their information are complementary with each other to formulate a more complete description about objects.

\noindent\textbf{Context Encoding with LSTM.} LSTM network shows its powerfulness in encoding the context information for image classification. In~\cite{lam2017fine}, Lam {\em et al.} address fine-grained image classification by mining informative image parts using a heuristic network, a successor network and a single layer LSTM. The heuristic network is responsible for extracting features from proposals and the successor network is responsible for predicting the new proposal offset. A single layer LSTM is used to fuse the information both for final object class prediction and also for the offset prediction. Attentional regions is discovered recurrently by incorporating a LSTM sub-network for multi-label image classification in \cite{wang2017multi}. The LSTM sub-network sequentially predict semantic labeling scores on the located regions and captures the spatial dependencies at the same time.

LSTM is used in our complementary part model to integrate the rich information hidden in different object proposals detected. Different from the single direction LSTM in~\cite{lam2017fine,wang2017multi}, we exploit a bi-directional LSTM to learn deep hierachical representation of all image patches. Experimental results show this strategy improve the performance substantially compared to the single layer LSTM.

\begin{figure*}[ht]
  \centering
  \includegraphics[width=1.0\linewidth]{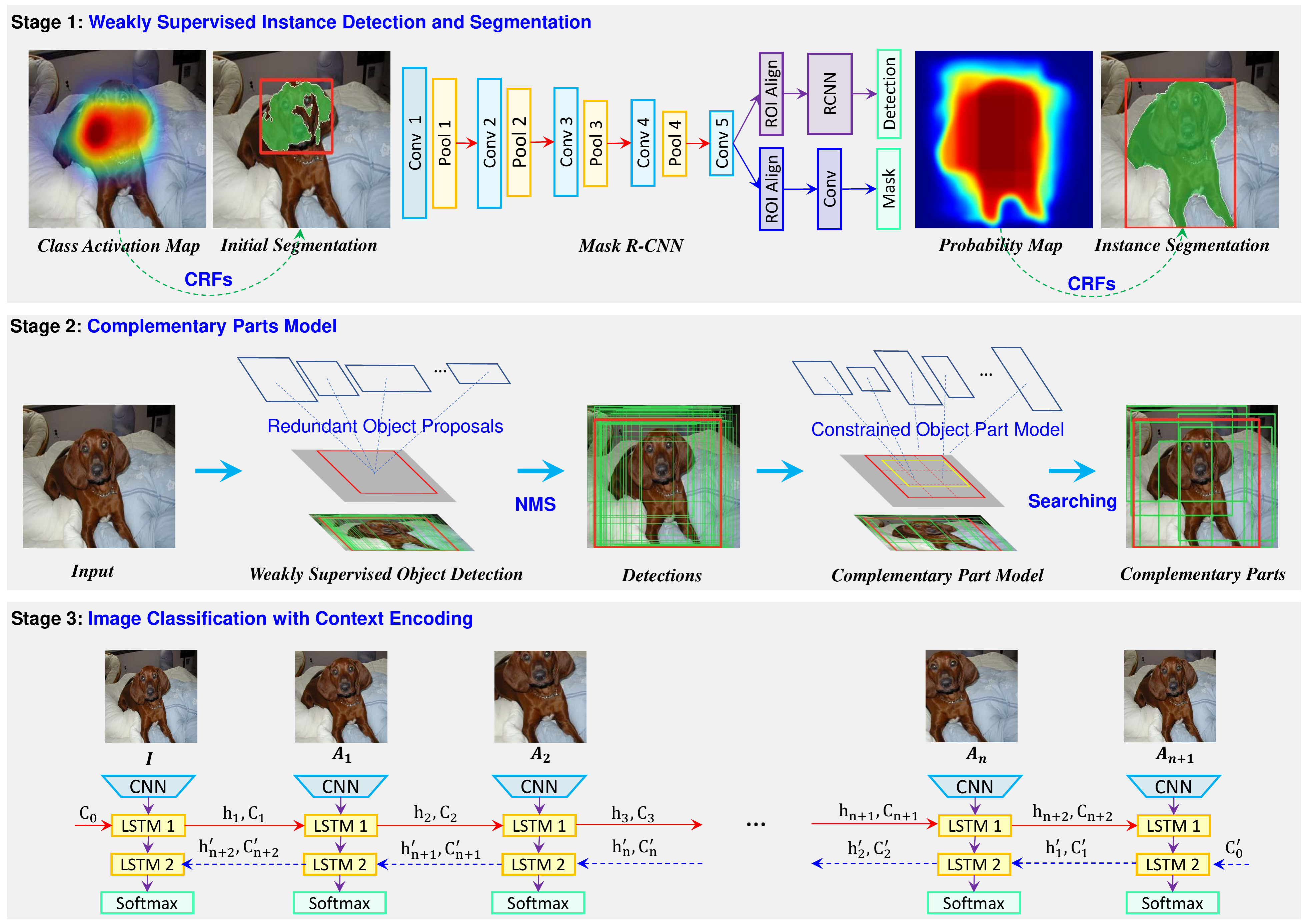}
  \caption{The proposed image classification pipeline based on weakly supervised complementary parts model. From top to bottom: (a) Weakly Supervised Object Detection and Instance Segmentation: The first step initializes the segmentation probability map by CAM~\cite{zhou2016learning}, and obtaining coarse instance segmentation maps by CRF~\cite{sutton2012introduction}. Then the segments and bounding boxes are used as groundtruth annotations for training Mask R-CNN~\cite{he2017mask} in an iterative manner. (b) Complementary Parts Model: Search for complementary object proposals to form the object parts model. (c) Image Classification with Context Encoding: Two LSTMs~\cite{hochreiter1997long} are stacked together to fuse and encode the partial information provided by different object parts.}
  \label{Fig:WSCPM}
\end{figure*}

\section{Weakly Supervised Complementary Parts Model}

\subsection{Overview}
Given an image $\boldsymbol{I}$ and its corresponding image label $\boldsymbol{c}$, the method proposed in this paper aims to mine discriminative parts $\mathcal{M}$ of an object that capture complementary information via object detection and then fuse the mined complementary parts for image classification. This is a reversal of a current trend~\cite{he2017mask, ren2015faster, lin2017feature}, which fine-tunes image classification models for object detection. Since we do not have labeled part locations but image level labels only, we formulate our problem in a weakly supervised manner. We adopt an iterative refinement pipeline to improve the estimation of object parts. Then we build a classifier utilizing the rich context representation focusing on object parts to boost classification performance. We decompose our pipeline into three stages, as shown in Fig~\ref{Fig:WSCPM}, namely, weakly supervised object detection and instance segmentation, complementary part model mining and image classification with context encoding.

\subsection{Weakly Supervised Object Detection and Instance Segmentation}
\noindent\textbf{Coarse Object Mask Initialization.}
Given an image $\boldsymbol{I}$ and its image label $\boldsymbol{c}$, the feature map of the last convolutional layer of a classification network is denoted as $\phi\left (\boldsymbol{I}, \theta \right ) \in {\mathbb{R}}^{K \times h \times  w}$, where $\theta$ represents the parameters of network $\phi$, $K$ is the number of channels, $h$ and $w$ are the height and width of the feature map respectively. Next, global average pooling is performed on $\phi$ to obtain the pooled feature $F_{k} = \sum_{x,y}\phi_{k}(x,y)$. The classification layer is added at the end and thus, the class activation map (CAM) for class $c$ is given as follows,
    \begin{equation}
    \begin{aligned}
		\boldsymbol{M}_{c}(x, y)=\sum_{k}w_{k}^{c}\phi_{k}(x,y),
    \end{aligned}
    \label{eq:class activation map}
    \end{equation}
where $w_{k}^{c}$ is the weight corresponding to class $c$ for the $k$-th channel in the global average pooling layer. The obtained class activation map $\boldsymbol{M}_{c}$ is upsampled to the original image size ${\mathbb{R}}^{ H \times  W}$ through bilinear interpolation. Since an image could have multiple object instances, multiple locally maximum responses could be observed on the class activation map $\boldsymbol{M}_c$. We apply multi-region level set segmentation~\cite{brox2006level} to this map to segment candidate object instances. Next, for each instance, we normalize the class activation to the range, $[0,1]$. Suppose we have $n$ object instances in CAM, we set up an object probability map $\boldsymbol{F} \in {\mathbb{R}}^{(n+1) \times H \times W}$ according to the normalized CAM. The first $n$ object probability maps denote the probability of a certain object existing in the image and the $(n + 1)$-th probability map represents the probability of the background. The background probability map is calculated as
    \begin{equation}
    \begin{aligned}
      \boldsymbol{F}^{n+1}_{ i \in{\mathbb{R}}^{ H \times  W}} = \max(1-\sum_{\iota =1}^{n}\boldsymbol{F}^{\iota}_{ i \in{\mathbb{R}}^{ H \times  W}}, 0).
    \end{aligned}
    \label{eq:bg map1}
    \end{equation}

Then a conditional random field (CRF)~\cite{sutton2012introduction} is used to extract higher-quality object instances. In order to apply CRFs, a label map $\boldsymbol{L}$ is generated according to the following formula,
    \begin{equation}
    \begin{aligned}
		\boldsymbol{L}_{i \in{\mathbb{R}}^{ H \times  W}} = \begin{cases}
		\lambda, \mathop{\arg\max}_{ \lambda }\boldsymbol{F}^{\lambda}_{i \in{\mathbb{R}}^{ H \times  W}} > \sigma_{c}
		\\
		0, otherwise
		\end{cases}
    \end{aligned}
    \label{eq:bg map1}
    \end{equation}
where $\sigma_{c}$ is always set to 0.8, a fixed threshold used to determine how certain a pixel belongs to an object or background. The label map $\boldsymbol{L}$ is then fed into a CRF to generate object instance segments, that are treated as pseudo groundtruth annotations for Mask-RCNN training. The parameters in the CRF are the same as in ~\cite{krahenbuhl2011efficient}. Fig~\ref{Fig:WSCPM} stage 1 shows the whole process of object instance segmentation.

\noindent\textbf{Jointly Detect and Segment Object Instances.} Given a set of segmented object instances, $\mathcal{S} = \left [ \mathcal{S}_1,\mathcal{S}_2,...\mathcal{S}_n \right ]$ of $\boldsymbol{I}$, and their corresponding class labels, generated in the previous stage, we obtain the minimum bounding box of each segment to form a set of proposals, $\mathcal{P} = \left [ \mathcal{P}_1,\mathcal{P}_2,...\mathcal{P}_n \right ]$. The proposals $\mathcal{P}$, segments $\mathcal{S}$ and their corresponding class labels are used for training Mask R-CNN for further proposal and mask refinement. In this way, we turn object detection and instance segmentation into fully supervised learning. We train Mask R-CNN with the same setting as in~\cite{he2017mask}.

\noindent\textbf{CRF-Based Segmentation.}
Suppose there are $m$ object proposals, $\mathcal{P}^{\star} = \left [ \mathcal{P}^{\star}_1,\mathcal{P}^{\star}_2,...,\mathcal{P}^{\star}_{m} \right ]$, and their corresponding segments, $\mathcal{S}^{\star} = \left [ \mathcal{S}^{\star}_1,\mathcal{S}^{\star}_2,...,\mathcal{S}^{\star}_{m} \right ]$ for image class $c$, whose classification score is above $\sigma_{0}$, a threshold used to remove outlier proposals. Then, a non-maximum suppression (NMS) procedure is applied to $m$ proposals with overlapping threshold $\tau$. Suppose $n$ object proposals remain afterwards, $\mathcal{O} = \left [ \mathcal{O}_{1},\mathcal{O}_{2},...,\mathcal{O}_{n} \right ]$, where $n \ll m$.

Most existing research utilizes NMS to suppress a large number of proposals sharing the same class label in order to obtain a small number of distinct object proposals. However, in our weakly supervised setting, proposals suppressed in the NMS process actually contain rich object parts information as shown in Fig~\ref{Fig:WSCPM}. Specifically, each proposal $\mathcal{P}^{\star}_{i} \in \mathcal{P}^{\star}$ suppressed by object proposal $\mathcal{O}_{j}$ can be considered as a complementary part of $\mathcal{O}_{j}$. Therefore, the suppressed proposals, $\mathcal{P}^{\star}_{i}$, can be used to further refine $\mathcal{O}_{j}$. We implement this idea by initializing a class probability map $\boldsymbol{F}^{\star} \in {\mathbb{R}}^{(n+1) \times H \times W}$. For each proposal $\mathcal{P}^{\star}_{i}$ suppressed by $\mathcal{O}_{j}$, we add the probability map of its proposal segmentation mask $\mathcal{S}^{\star}_{i}$ to the corresponding locations on $\boldsymbol{F}^{\star}_{j}$ by bilinear interpolation. The class probability map is then normalized to [0, 1]. For the $(n+1)$-th probability map for the background, it is defined as
    \begin{equation}
    \begin{aligned}
      \boldsymbol{F}^{\star, n+1}_{ i \in{\mathbb{R}}^{ H \times  W}} = \max(1-\sum_{\iota =1}^{n}\boldsymbol{F}^{\star, \iota}_{ i \in{\mathbb{R}}^{ H \times  W}}, 0).
    \end{aligned}
    \label{eq:bg map1}
    \end{equation}

Given the class probability maps $\boldsymbol{F}^{\star}$, CRF is applied again to refine and rectify instance segmentation results as described in the previous stage.

\noindent\textbf{Iterative Instance Refinement.} We alternate CRF-based segmentation and Mask R-CNN based detection and instance segmentation several times to gradually refine the localization and segmentation of object instances. Fig~\ref{Fig:WSCPM} shows the iterative instance refinement process.

\subsection{Complementary Parts Model}

\noindent\textbf{Model Definition.}
According to the analysis in the previous stage, given a detected object $\mathcal{O}_i$, its corresponding suppressed proposals, $\mathcal{P}^{\star, i} = \left [ \mathcal{P}^{\star, i}_1,\mathcal{P}^{\star, i}_2,...,\mathcal{P}^{\star, i}_{k} \right ]$, may contain useful object information and can localize correct object position. Then, it is necessary to identify the most informative proposals for the following classification task. In this section, we propose a complementary parts model $\mathcal{A}$ for image classification. This model is defined by a root part covering the entire object as well as its context, a center part covering the core region of the object and a fixed number of surrounding proposals that cover different object parts but still keep enough discriminative information.

A complementary parts model for an object with $n$ parts is defined as a $(n+1)$-tuple $\mathcal{A} = \left [ \boldsymbol{A}_1, ..., \boldsymbol{A}_n,\boldsymbol{A}_{n+1}  \right ] $, where $\boldsymbol{A}_1$ is the object center part, $\boldsymbol{A}_{n+1}$ is the root part, and $\boldsymbol{A}_i$ is the $i$-th part. Each part model is defined by a tuple $\boldsymbol{A}_i = \left [ \phi_i, \boldsymbol{u}_i \right ]$, where $\phi_i$ is the feature of the $i$-th part, $\boldsymbol{u}_i$ is a $\mathbb{R}^{4}$ dimensional tuple that describes the geometric information of a part, namely part center and part size $(x_i, y_i, w_i, h_i)$. A potential parts model without any missing parts is called an object hypothesis. To make object parts complementary to each other, the differences in their appearance features or locations should be as large as possible while the combination of parts scores should also be as large as possible. Such criteria serve as constraints during the search for discriminative parts that are complementary to each other. The score $\mathcal{S}\left ( \mathcal{A} \right )$ of an object hypothesis is given by the summed score of all object parts minus appearance similarities and spatial overlap between different parts.
    \begin{equation}
    \begin{aligned}
      \mathcal{S}\left ( \mathcal{A} \right ) = & \sum _{\iota=1}^{n+1}f\left ( \phi_{\iota} \right ) \\
       & - \lambda_0 \sum _{p=1}^{n}\sum _{q=p+1}^{n+1}\left [ d_s(\phi_p,\phi_q) + \beta_0 IoU(\boldsymbol{u}_p,\boldsymbol{u}_q) \right ],
    \end{aligned}
    \label{eq:hypothesis cost}
    \end{equation}
where $f\left ( \phi_k \right )$ is the score of the $k$-th part in the classification branch of Mask R-CNN, $d_s(\phi_p,\phi_q) = \left \|  \phi_p - \phi_q \right \| ^2$ is the semantic similarity and $IoU(\boldsymbol{u}_p,\boldsymbol{u}_q)$ is the spatial overlap between parts $p$ and $q$, and there are two constant parameters $\lambda_0=0.01$ and $\beta_0=0.1$. Given a set of object hypotheses, we can choose a hypothesis that achieves the maximum score as the final object part model. Searching for the optimal subset of proposals maximizing the above score is a combinatorial optimization problem, which is computationally expensive. In the following, we seek an approximate solution using a fast heuristic algorithm.

\noindent\textbf{Part Location Initialization.} To initialize a parts model, we simplify part estimation by designing a grid-based object parts template that follows two basic rules. First, every part should contain enough discriminative information; Second, the differences between part pairs should be as large as possible. As shown in Fig~\ref{Fig:WSCPM}, deep convolutional neural networks have demonstrated its ability in localizing the most discriminative parts of an object. Thus, we set the root part $\boldsymbol{A}_{n+1}$ to be the object proposal $\mathcal{O}_i$ that represents the entire object. Then, a $s \times s (= n)$ grid centered at $\boldsymbol{A}_{n+1}$ is created. The size of each grid cell is $\frac{w_{n+1}}{s} \times \frac{h_{n+1}}{s}$, where $w_{n+1}$ and $h_{n+1}$ are the width and height of the root part $\boldsymbol{A}_{n+1}$. The center grid cell is assigned to the object center part. The rest of the grid cells are assigned to part $\boldsymbol{A}_i$, where $i \in [2, 3, ..., n]$. Then, we initialize each part $\boldsymbol{A}_{i} \in \boldsymbol{A}$ to be the proposal $\mathcal{P}^{\star}_{j} \in \mathcal{P}^{\star}$ closest to the assigned grid cell.

\noindent\textbf{Parts Model Search.} For a model with $n$ object parts (we exclude the $(n+1)$-th part as it is a root part) and $k$ candidate suppressed proposals, the objective function is defined as
    \begin{equation}
    \begin{aligned}
      \widehat{\mathcal{A}} = \mathop{\arg\max}_{ \mathcal{A} \in \mathcal{S}_{\mathcal{A}}} \mathcal{S}\left ( \mathcal{A} \right ),
    \end{aligned}
    \label{eq:hypothesis cost}
    \end{equation}
where $K = C_{k}^{n}, k \gg n$ is the total number of object hypothesises,  $\mathcal{S}_{\mathcal{A}} = \left [ \mathcal{A}^1,\mathcal{A}^1,...,\mathcal{A}^K \right ]$
is the set of object hypotheses. As mentioned earlier, directly searching for an optimal parts model can be intractable. Thus, we adopt a greedy search strategy to search for $\widehat{\mathcal{A}}$. Specifically, we sequentially go through every $\boldsymbol{A}_{i}$ in $\boldsymbol{A}$ and find the optimal object part for $\boldsymbol{A}_{i}$ in $\mathcal{P}^{\star}$ that minimizes $\widehat{\mathcal{A}}$. The overall time complexity is reduced from exponential to linear ($O(nk)$). In Fig~\ref{Fig:WSCPM}, we can see that the object hypotheses generated during the search process cover different parts of the object and do not focus on the core region only.

\begin{figure}[ht]
  \centering
  \includegraphics[width=1.0\linewidth]{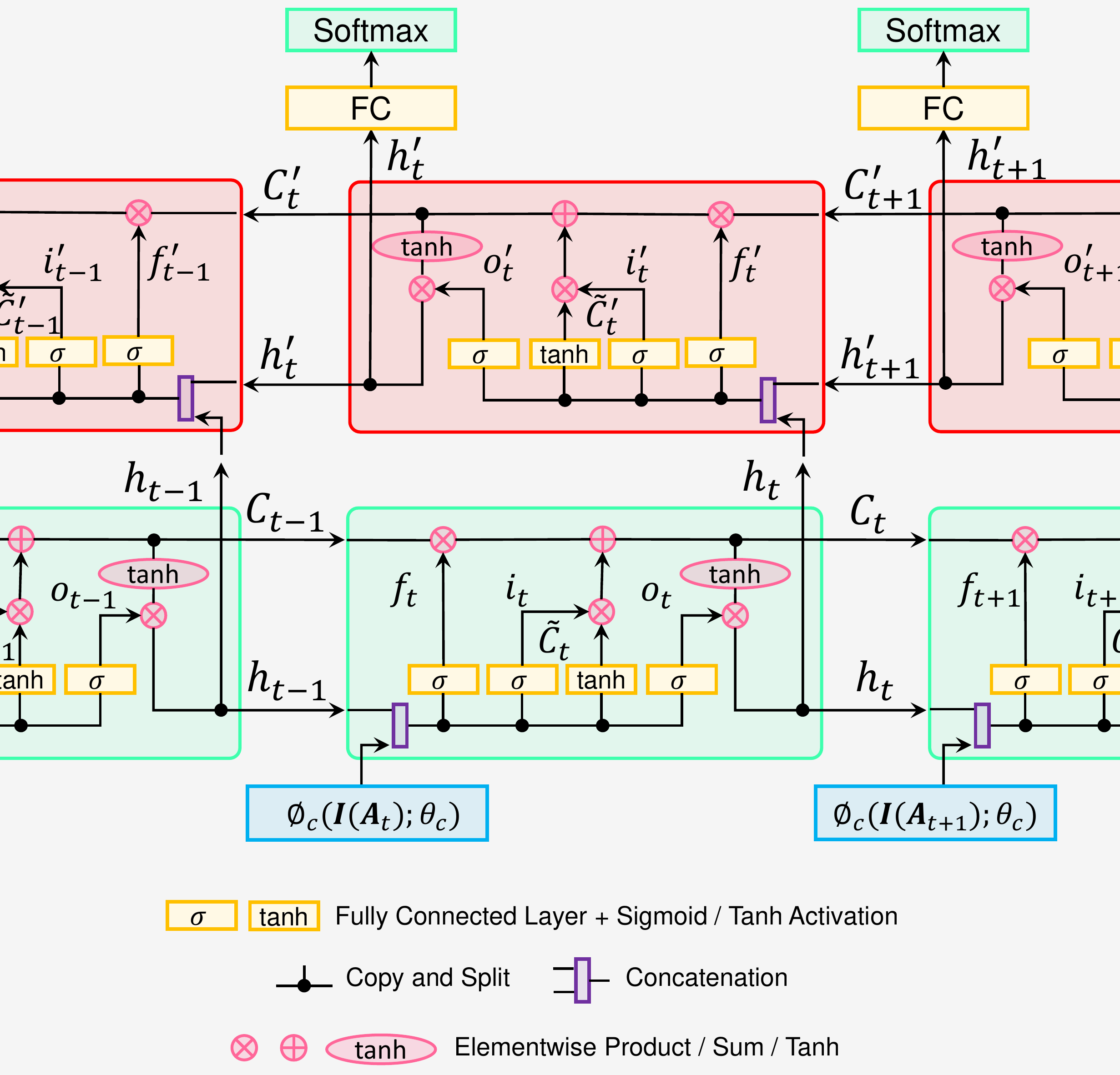}
  \caption{Context encoded image classification based on LSTMs. Two standard LSTMs~\cite{hochreiter1997long} are stacked together. They have opposite scanning orders.}
  \label{Fig:LSTM}
\end{figure}

\subsection{Image Classification with Context Encoding}
\noindent\textbf{CNN Feature Extractor Fine-tuning.} Given an input image $\boldsymbol{I}$ and the parts model $\mathcal{A} = \left [ \boldsymbol{A}_1, ..., \boldsymbol{A}_n,\boldsymbol{A}_{n+1}  \right ]$ constructed in the previous stage, the image patches corresponding to the parts are denoted as $\boldsymbol{I}\left ( \mathcal{A} \right ) = \left [ \boldsymbol{I}\left ( \boldsymbol{A}_1 \right ) ,\boldsymbol{I}\left ( \boldsymbol{A}_2 \right ), ..., \boldsymbol{I}\left ( \boldsymbol{A}_n \right ), \boldsymbol{I}\left ( \boldsymbol{A}_{n+1} \right ) \right ] $. During image classification, random crops of images are often used to train the model. Thus, apart from the $(n+1)$ patches, we append a random crop of the original image as the $(n+2)$-nd image patch. The motivation for adding a randomly cropped patch is to include more context information during training since those patches corresponding to object parts primarily focus on the object itself. Every patch shares the same label with the original image it is cropped from. All patches from all the original training images form a new training set, which is used to fine-tune a CNN model pretrained on ImageNet. This fine-tuned model serves as the feature extractor for all image patches.

\noindent\textbf{Stacked LSTM for Feature Fusion.} Here we propose a stacked LSTM module $\phi_l\left ( \cdot ; \theta_l \right )$ for feature fusion and performance boosting, which is shown in Fig~\ref{Fig:LSTM}. First, the $(n+2)$ patches from a complementary parts model are fed through the CNN feature extractor $\phi_c\left ( \cdot;\theta_c \right )$ trained in the previous step. The output from this step is denoted as $\Psi \left ( \boldsymbol{I} \right ) = \left [ \phi_c\left ( \boldsymbol{I};\theta_c \right ),\phi_c\left ( \boldsymbol{I}\left ( \boldsymbol{A}_1 \right );\theta_c \right ),... ,\phi_c\left ( \boldsymbol{I}\left ( \boldsymbol{A}_{n+2} \right );\theta_c \right ) \right ]$. Next, we build a two-layer stacked LSTM to fuse the extracted features $\Psi \left ( \boldsymbol{I} \right )$. The hidden state of the first LSTM is fed into the second LSTM layer, but the second LSTM follows the reversed order of the first one. Let $D (= 256)$ be the dimension of the hidden state. We use softmax to generate the class probability vector for each part $\boldsymbol{A}_{i}$, $f\left ( \phi_l\left ( \boldsymbol{I}\left ( \boldsymbol{A}_{i} \right ) ; \theta_l \right ) \right ) \in {\mathbb{R}}^{\mathcal{C} \times  1}$. The loss function for final image classification is defined as follows,
    \begin{equation}
    \begin{aligned}
      \mathcal{L}(\boldsymbol{I},\boldsymbol{y}_I) = &-\sum_{k=1}^{\mathcal{C}} y^k \log f ^k\left ( \phi_l\left ( \boldsymbol{I} ; \theta_l \right ) \right ) \\
      &- \sum_{i=1}^{n+2} \sum_{k=1}^{\mathcal{C}} \gamma_i y^k \log f ^k\left ( \phi_l\left ( \boldsymbol{I}\left ( \boldsymbol{A}_{i} \right ) ; \theta_l \right ) \right ),
    \end{aligned}
    \label{eq:lstm cost}
    \end{equation}
where $f^k\left ( \phi_l\left ( \boldsymbol{I} ; \theta_l \right ) \right )$ is the probability that image $\boldsymbol{I}$ belongs to the $k$-th class, $f ^k\left ( \phi_l\left ( \boldsymbol{I}\left ( \boldsymbol{A}_{i} \right ) ; \theta_l \right ) \right )$ is the probability that image patch $\boldsymbol{I}\left ( \boldsymbol{A}_{i} \right )$ belongs to the $k$-th class, and $\gamma_i$ is a constant weight for the $i$-th patch. Here we have two settings: first, the single loss sets $\gamma _i = 0\left ( i=2,...,n+2 \right )$, and keeps only one loss at the start of the sequence; second, the multiple losses sets $\gamma _i = 1\left ( i=2,...,n+2 \right )$. Experimental results indicate that, in comparison to a single loss for the last output from the second LSTM, multiple losses used here improve classification accuracy by a significant margin.

\begin{figure*}[ht]
  \centering
  \includegraphics[width=0.9\linewidth]{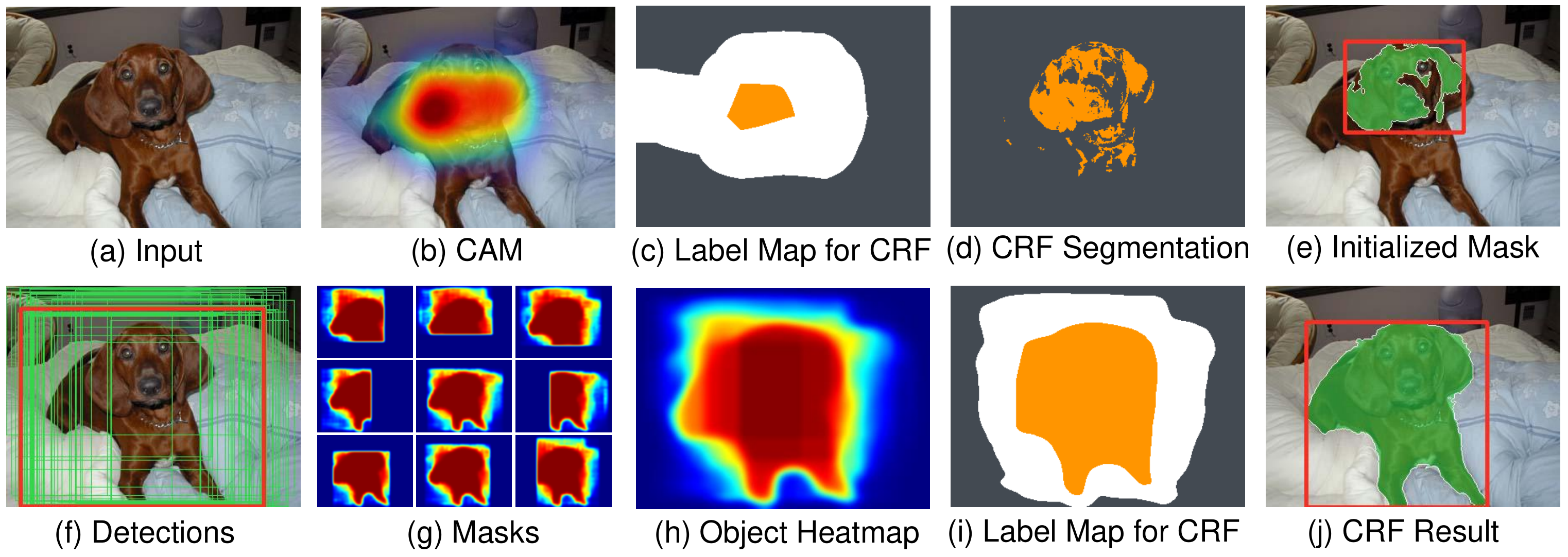}
  \caption{Example intermediate results for training Mask R-CNN. \textit{First row}: pseudo object mask and object bounding box are generated with CAM and CRF refinement. \textit{Second row}: With previous pseudo groundtruth generated, object mask and object bounding box are further refined with Mask R-CNN.}
  \label{Fig:ExampleResult}
\end{figure*}

\section{Experimental Results}
\subsection{Implementation Details}\label{sec:implementation}
All experiments have been conducted on NVIDIA TITAN X(Maxwell) GPUs with 12GB memory using Caffe~\cite{jia2014caffe}. No annotated parts are used. $n$ is set to 9 for all experiments.

In the mask initialization stage, we fine-tune from ImageNet pre-trained GoogleNet with batch normalization~\cite{Sergey2015batchnorm} on target datasets. 
The initial learning rate is 0.001 and is divided by 10 after every 40000 iterations with the standard SGD optimizer. Training converges after 70000 iterations.
In the Mask R-CNN refinement process, we adopt ResNet-50 with Feature Pyramid Network (FPN) as the backbone and pre-train the network on the COCO dataset following the same setting described in \cite{he2017mask}. We then fine-tune the model on our target datasets. During training, image-centric training is used and the input images are resized such that their shorter side is 800 pixels. Each mini-batch contains 1 image per GPU and each image has 512 sampled ROIs. The model is trained on 4 GPUs for 150k iterations with an initial learning rate 0.001, which is divided by 10 at 120k iterations. We use the standard SGD optimizer and a weight decay of 0.0001. The momentum is set to 0.9. Unless specified, the settings we use for different algorithms follow their original settings respectively~\cite{zhou2016learning, Szegedy2016googlenet, brox2006level, krahenbuhl2011efficient, he2017mask}. Example intermediate results of Mask R-CNN training are shown in Fig~\ref{Fig:ExampleResult}.


For the last stage, we adopt GoogleNet with batch normalization~\cite{Sergey2015batchnorm} as the backbone network for Stanford Dogs 120 and Caltech-UCSD Birds 2011-200 datasets and the Caltech256 dataset. 
First, we fine-tune the pretrained network on the target dataset with the generated object parts. The parameters are the same as those used in the first stage. 
Next, we build a Stacked LSTM module and treat the features of the $n+2$ image patches as training sequences. 
We train the model with 4 GPUs and set the learning rate to 0.001, which is decreased by a factor of 10 for every 8000 iterations. We adopt the standard SGD optimizer, momentum is set to 0.9, and the weight decay is 0.0002. Training converges at 16000 iterations.


\subsection{Fine-grained Image Classification}
\noindent\textbf{Stanford Dogs 120.} Stanford Dogs 120 contains 120 categories of dogs. There are 12000 images for training, and 8580 images for testing. The training procedure follows the steps described in Section~\ref{sec:implementation}. 

To perform fair comparisons with existing state-of-the-art algorithms, we divide our experiments into two groups. The first group consists of algorithms that use the original training data only and the second group is composed of methods that use extra training data. In each group, we set our baseline accordingly. In the first group, we directly fine-tune the GoogleNet pretrained on ImageNet with the input image size set to 448 x 448, which is adopted by other algorithms~\cite{jianf17, Liu16FCAN, sun18} in the comparison and the classification accuracy achieved is $85.2\%$. This serves as our baseline model and we then add the proposed stacked LSTM over a complementary parts model. Our stacked LSTM is trained with both single loss and multiple losses, which achieves a classification accuracy of $92.4\%$ and $93.9\%$ respectively. Both of our proposed variants outerperform existing state-of-the-art by a clear margin. In the second group, we perform selective joint fine-tuning (SJFT) with images retrieved from ImageNet, and the input image size is set to 224 x 224 to obtain our baseline network. The classification accuracy of our baseline is $92.1\%$, $1.8\%$ higher than the SJFT with ResNet-152 counterpart. With our stacked LSTM plugged in and trained with both single loss and multiple losses, the performance is further boosted to $96.3\%$ and $97.1\%$ respectively, surpassing the current state of the art by $6\%$ and $6.8\%$. These experimental results suggest that our proposed pipeline is superior than all existing algorithms. It is worth noting that the method in \cite{krause2015unreasonable} is not directly comparable to ours because it uses a large amount of extra training data from the Internet in addition to ImageNet.

\begin{table}[t]\small
\setlength{\abovecaptionskip}{10pt}
\setlength{\belowcaptionskip}{-10pt}
\begin{center}
\begin{tabular}{@{}lccc@{}}
\toprule
Method                                             & Accuracy(\%)       \\ \midrule
MAMC \cite{sun18}                                  & 85.2               \\
Inception-v3 \cite{krause2015unreasonable}         & 85.9               \\
RA-CNN \cite{jianf17}                              & 87.3               \\
FCAN \cite{Liu16FCAN}                              & 88.9               \\
GoogleNet (our baseline)                               & 85.2               \\
baseline + Feature Concatenation                   & 88.1               \\
baseline + Multiple Average                        & 85.2               \\
baseline + Stacked LSTM + Single Loss              & 92.4               \\
baseline + Stacked LSTM + Multi-Loss (default)     & \textbf{93.9}      \\\midrule
Web Data + Original Data \cite{krause2015unreasonable} & 85.9             \\
SJFT with ResNet-152 \cite{GeYu17}                 & 90.3               \\
SJFT with GoogleNet (our baseline)                     & 92.1               \\
baseline + Feature Concatenation                   & 93.2               \\
baseline + Multiple Average                        & 92.2               \\
baseline + Stacked LSTM + Single Loss             & 96.3               \\
baseline + Stacked LSTM + Multi-Loss (default)    & \textbf{97.1}      \\
\bottomrule
\end{tabular}
\end{center}
\caption{Classification results on Stanford Dogs 120. Two sections are divided by the horizontal separators, namely (from top to bottom) Experiments without SJFT and Experiments with SJFT.}
\label{tab:StanfordDogs120}
\end{table}

\noindent\textbf{Caltech-UCSD Birds 2011-200.}
Caltech-UCSD Birds 2011-200 (CUB200) consists of 200 bird categories. 5994 images are used for training, and 5794 images for testing. 

Our experiments here are split into two groups. In the first group, no extra training data is used. Our baseline model in this group is a directly fine-tuned GoogleNet model that achieves a classification accuracy of $82.6\%$. We then add the Stacked LSTM module and train the model with both single loss and multiple losses, which achieves a classification accuracy of $87.6\%$ and $90.3\%$ respectively, outperforming all other algorithms in this comparison~\cite{zheng17, Yu_2018_ECCV, wang16, Lam_17_HSNet}. Compared to HSNet, our model does not use any parts annotations in the training stage while HSNet is trained with groundtruth parts annotations. In the second group, our baseline model still uses GoogleNet as the backbone and performs SJFT with images retrieved from ImageNet. It achieves a classification accuracy of $82.8\%$. By adding the Stacked LSTM module, the accuracy of the model trained with single loss is $87.7\%$ and the model trained with multiple losses is $90.4\%$. When the top performing result in the first group is compared to that of the second group, it can be concluded that SJFT contributes little to the performance gain ($0.1\%$ gains) and our proposed method is effective and solid, contributing much to the final performance ($7.7\%$ higher than the baseline). It is worth noting that, in \cite{cui18}, a subset of ImageNet and iNaturalist~\cite{van17} most similar to CUB200 are used for training, and in \cite{krause2015unreasonable}, a large amount of web data are also used in the training phase.

\begin{table}[t]\small
\setlength{\abovecaptionskip}{10pt}
\setlength{\belowcaptionskip}{-10pt}
\begin{center}
\begin{tabular}{@{}lccc@{}}
\toprule
Method                                             & Accuracy(\%)       \\ \midrule
MACNN \cite{zheng17}                               & 86.5               \\
HBP \cite{Yu_2018_ECCV}                            & 87.2               \\
DFB \cite{wang16}                                  & 87.4               \\
HSNet \cite{Lam_17_HSNet}                          & 87.5               \\
GoogleNet (our baseline)                           & 82.6               \\
baseline + Stacked LSTM + Single Loss              & 87.6               \\
baseline + Stacked LSTM + Multi-Loss               & \textbf{90.3}               \\\midrule
ImageNet + iNat Finetuning \cite{cui18}            & 89.6               \\
SJFT with GoogleNet (our baseline)                 & 82.8               \\
baseline + Stacked LSTM + Single Loss              & 87.7               \\
baseline + Stacked LSTM + Multi-Loss               & \textbf{90.4}              \\
\bottomrule
\end{tabular}
\end{center}
\caption{Classification results on CUB200. Two sections are divided by the horizontal separators, namely (from top to bottom) Experiments without SJFT and Experiments with SJFT.}
\label{tab:CUB200}
\end{table}

\begin{table}[t]\small
\setlength{\abovecaptionskip}{10pt}
\setlength{\belowcaptionskip}{-10pt}
\begin{center}
\begin{tabular}{@{}lccc@{}}
\toprule
Method                                               & Accuracy(\%)          \\ \midrule
ZF Net \cite{zfnet}                                  & 74.2$\pm$0.3          \\
VGG-19 + VGG-16 \cite{vgg}                           & 86.2$\pm$0.3          \\
VGG-19 + GoogleNet +AlexNet \cite{kim15}             & 86.1                  \\
$L^{2}$-SP \cite{Li2018ExplicitIB}                   & 87.9$\pm$0.2          \\
GoogleNet (our baseline)                             & 84.1$\pm$0.2               \\
baseline + Stacked LSTM + Single-Loss                & 90.1$\pm$0.2               \\
baseline + Stacked LSTM + Multi-Loss                 & \textbf{93.5$\pm$0.2}      \\\midrule
SJFT with ResNet-152 \cite{GeYu17}                   & 89.1$\pm$0.2          \\
SJFT with GoogleNet (our baseline)                       & 86.3$\pm$0.2          \\
baseline + Stacked LSTM + Single-Loss                & 90.1$\pm$0.2          \\
baseline + Stacked LSTM + Multi-Loss                 & \textbf{94.3$\pm$0.2} \\
\bottomrule
\end{tabular}
\end{center}
\caption{Classification results on Caltech 256. Two sections are divided by the horizontal separators, namely (from top to bottom) Experiments without SJFT and Experiments with SJFT.}
\label{tab:Caltech256}
\end{table}


\subsection{Generic Object Recognition}
\noindent\textbf{Caltech 256.} There are 256 object categories and 1 background cluster class in Caltech 256. A minimum number of 80 images per category are provided for training, validation and testing. As a convention, results are reported with the number of training samples per category falling between 5 and 60. We follow the same convention and report the result with the number of training sample per category set to 60. In this experiment, GoogleNet is adopted as our backbone network and the input image size is 224 x 224. We train our model with mini-batch size set to 8 on each GPU.


In Table~\ref{tab:Caltech256}, as described previously, we conduct our experiments under two settings. For the first setting, no extra training data is used. We fine-tune the pretrained GoogleNet on the target dataset and treat the fine-tuned model as our baseline model, which achieves a classification accuracy of $84.1\%$. By adding our proposed Stacked LSTM module, the accuracy is increased by a large margin to $90.1\%$ for Single Loss and to $93.5\%$ for multiple losses respectively, outerperforming all methods listed in the table. Also, it is $4.1\%$ higher than its ResNet-152 counterpart. For the second setting, we adopt SJFT ~\cite{GeYu17} with GoogleNet as our baseline model, which achieves a classification accuracy of $86.3\%$. Then we add our proposed Stacked LSTM module and the final performance is increased by $3.8\%$ for single loss and $8.0\%$ for multiple losses. Our method with GoogleNet as backbone network outerperfoms current state-of-the-art by $5.2\%$, demonstrating that our proposed algorithm is solid and effective.

\subsection{Ablation Study}
\textbf{Ablation Study on Complementary Parts Mining.} The ablation study is performed on the CUB200 dataset with GoogleNet as the backbone network. The classification accuracy of our reference model with $n=9$ parts on this dataset is $90.3\%$. First, when the number of parts $n$ is set to $2$, $4$, $6$, $9$, $12$, $16$, and $20$ in our model, the corresponding classification accuracy is respectively $85.3\%$, $87.9\%$, $89.1\%$, $90.3\%$, $87.6\%$, $86.8\%$ and $85.9\%$.
Obviously the best result is achieved when $n=9$.
Second, if we use object features only in our reference model, the classification accuracy drops to $90.0\%$. Third, if we use image features only, the performance drops to $82.8\%$.
Fourth, if we simply use the uniform grid cells as the object parts without further optimization, the performance drops to $78.3\%$, which indicates our search for the best parts model plays an important role in escalating the performance.
Fifth, instead of a grid-based object parts initialization, we randomly sample $n=9$ suppressed object proposals around the bounding box of the surviving proposal, and the performance drops to $86.9\%$. Lastly, we discover that the part order in LSTM does not matter. We randomly shuffle the part order during training and testing, and the classification accuracy remains the same.

\textbf{Ablation Study on Context Fusion.}We perform an ablation study on Stanford Dogs 120 for the context fusion stage. We first replace the multiple losses with the single loss and the accuracy drops from $93.9\%$ to $92.4\%$. This suggests that multiple losses help regularize the training process and produce more discriminative features for image classification. We then keep the multiple losses setting in subsequent experiments. Second, the Stacked LSTM module is removed and we conduct experiments with two settings, a feature concatenation module and an averaging module. In the feature concatenation module, the features of all the $n+2$ parts are concatenated. And in the averaging module, the classification output of multiple features are summed. The classification accuracies achieved are decreased by $5.8\%$ and $8.7\%$ respectively. The performance drop suggests that fusing $n+2$ image patches through LSTM is helpful for final image classification.

\subsection{Inference Time Complexity.}
The inference time of our implementation is summarised as follows: in the complementary parts model search phase, the time for processing an image with its shorter edge set to $800$ pixels is around $277ms$; in the context encoding phase, the running time on an image of size $448 \times 448$ is about $63ms$, and on an image of size $224 \times 224$ is about $27ms$.
\vspace{0mm}

\section{Conclusions}
In this paper, we have presented a new pipeline for fine-grained image classification, which is based on a complementary part model. Different from previous work which focuses on learning the most discriminative parts for image classification, our scheme mines complementary parts that contain partial object descriptions in a weakly supervised manner. After getting object parts that contain rich information, we fuse all the mined partial object descriptions with bi-directional stacked LSTM to encode these complementary information for classification. Experimental results indicate that the proposed method is effective and outperforms existing state-of-the-art by a large margin. Nevertheless, how to build the complementary part model in a more efficient and accurate way remains an open problem for further investigation.
{\small
\bibliographystyle{ieee}
\bibliography{cpm}

\begin{thebibliography}{10}\itemsep=-1pt

\bibitem{bach2015pixel}
S.~Bach, A.~Binder, G.~Montavon, F.~Klauschen, K.-R. M{\"u}ller, and W.~Samek.
\newblock On pixel-wise explanations for non-linear classifier decisions by
  layer-wise relevance propagation.
\newblock {\em PloS one}, 10(7):e0130140, 2015.

\bibitem{bau2017network}
D.~Bau, B.~Zhou, A.~Khosla, A.~Oliva, and A.~Torralba.
\newblock Network dissection: Quantifying interpretability of deep visual
  representations.
\newblock In {\em 2017 IEEE Conference on Computer Vision and Pattern
  Recognition (CVPR)}, pages 3319--3327. IEEE, 2017.

\bibitem{brox2006level}
T.~Brox and J.~Weickert.
\newblock Level set segmentation with multiple regions.
\newblock {\em IEEE Transactions on Image Processing}, 15(10):3213--3218, 2006.

\bibitem{cui18}
Y.~Cui, Y.~Song, C.~Sun, A.~Howard, and S.~J. Belongie.
\newblock Large scale fine-grained categorization and domain-specific transfer
  learning.
\newblock {\em CoRR}, abs/1806.06193, 2018.

\bibitem{diba2016weakly}
A.~Diba, V.~Sharma, A.~M. Pazandeh, H.~Pirsiavash, and L.~V. Gool.
\newblock Weakly supervised cascaded convolutional networks.
\newblock In {\em 2017 {IEEE} Conference on Computer Vision and Pattern
  Recognition, {CVPR} 2017, Honolulu, HI, USA, July 21-26, 2017}, pages
  5131--5139, 2017.

\bibitem{durand2017wildcat}
T.~Durand, T.~Mordan, N.~Thome, and M.~Cord.
\newblock Wildcat: Weakly supervised learning of deep convnets for image
  classification, pointwise localization and segmentation.
\newblock In {\em IEEE Conference on Computer Vision and Pattern Recognition
  (CVPR 2017)}, volume~2, 2017.

\bibitem{durand2016weldon}
T.~Durand, N.~Thome, and M.~Cord.
\newblock Weldon: Weakly supervised learning of deep convolutional neural
  networks.
\newblock In {\em Proceedings of the IEEE Conference on Computer Vision and
  Pattern Recognition}, pages 4743--4752, 2016.

\bibitem{Dvornik_2017_ICCV}
N.~Dvornik, K.~Shmelkov, J.~Mairal, and C.~Schmid.
\newblock Blitznet: A real-time deep network for scene understanding.
\newblock In {\em The IEEE International Conference on Computer Vision (ICCV)},
  Oct 2017.

\bibitem{everingham2010pascal}
M.~Everingham, L.~Van~Gool, C.~K. Williams, J.~Winn, and A.~Zisserman.
\newblock The pascal visual object classes (voc) challenge.
\newblock {\em International journal of computer vision}, 88(2):303--338, 2010.

\bibitem{felzenszwalb2008discriminatively}
P.~Felzenszwalb, D.~McAllester, and D.~Ramanan.
\newblock A discriminatively trained, multiscale, deformable part model.
\newblock In {\em Computer Vision and Pattern Recognition, 2008. CVPR 2008.
  IEEE Conference on}, pages 1--8. IEEE, 2008.

\bibitem{jianf17}
J.~Fu, H.~Zheng, and T.~Mei.
\newblock Look closer to see better: Recurrent attention convolutional neural
  network for fine-grained image recognition.
\newblock In {\em 2017 {IEEE} Conference on Computer Vision and Pattern
  Recognition, {CVPR} 2017, Honolulu, HI, USA, July 21-26, 2017}, pages
  4476--4484, 2017.

\bibitem{Ge_2018_CVPR}
W.~Ge, S.~Yang, and Y.~Yu.
\newblock Multi-evidence filtering and fusion for multi-label classification,
  object detection and semantic segmentation based on weakly supervised
  learning.
\newblock In {\em The IEEE Conference on Computer Vision and Pattern
  Recognition (CVPR)}, June 2018.

\bibitem{GeYu17}
W.~Ge and Y.~Yu.
\newblock Borrowing treasures from the wealthy: Deep transfer learning through
  selective joint fine-tuning.
\newblock In {\em IEEE Conference on Computer Vision and Pattern Recognition
  (CVPR)}, July 2017.

\bibitem{girshick2014rich}
R.~Girshick, J.~Donahue, T.~Darrell, and J.~Malik.
\newblock Rich feature hierarchies for accurate object detection and semantic
  segmentation.
\newblock In {\em Proceedings of the IEEE conference on computer vision and
  pattern recognition}, pages 580--587, 2014.

\bibitem{griffin2007caltech}
G.~Griffin, A.~Holub, and P.~Perona.
\newblock Caltech-256 object category dataset.
\newblock 2007.

\bibitem{he2017mask}
K.~He, G.~Gkioxari, P.~Doll{\'a}r, and R.~Girshick.
\newblock Mask r-cnn.
\newblock In {\em Computer Vision (ICCV), 2017 IEEE International Conference
  on}, pages 2980--2988. IEEE, 2017.

\bibitem{he2016deep}
K.~He, X.~Zhang, S.~Ren, and J.~Sun.
\newblock Deep residual learning for image recognition.
\newblock In {\em Proceedings of the IEEE conference on computer vision and
  pattern recognition}, pages 770--778, 2016.

\bibitem{hochreiter1997long}
S.~Hochreiter and J.~Schmidhuber.
\newblock Long short-term memory.
\newblock {\em Neural computation}, 9(8):1735--1780, 1997.

\bibitem{Sergey2015batchnorm}
S.~Ioffe and C.~Szegedy.
\newblock Batch normalization: Accelerating deep network training by reducing
  internal covariate shift.
\newblock pages 448--456, 2015.

\bibitem{jia2014caffe}
Y.~Jia, E.~Shelhamer, J.~Donahue, S.~Karayev, J.~Long, R.~Girshick,
  S.~Guadarrama, and T.~Darrell.
\newblock Caffe: Convolutional architecture for fast feature embedding.
\newblock In {\em Proceedings of the 22nd ACM international conference on
  Multimedia}, pages 675--678. ACM, 2014.

\bibitem{KhoslaYaoJayadevaprakashFeiFei_FGVC2011}
A.~Khosla, N.~Jayadevaprakash, B.~Yao, and L.~Fei-Fei.
\newblock Novel dataset for fine-grained image categorization.
\newblock In {\em First Workshop on Fine-Grained Visual Categorization, IEEE
  Conference on Computer Vision and Pattern Recognition}, Colorado Springs, CO,
  June 2011.

\bibitem{kim15}
Y.~Kim, T.~Jang, B.~Han, and S.~Choi.
\newblock Learning to select pre-trained deep representations with bayesian
  evidence framework.
\newblock In {\em 2016 {IEEE} Conference on Computer Vision and Pattern
  Recognition, {CVPR} 2016, Las Vegas, NV, USA, June 27-30, 2016}, pages
  5318--5326, 2016.

\bibitem{krahenbuhl2011efficient}
P.~Kr\"{a}henb\"{u}hl and V.~Koltun.
\newblock Efficient inference in fully connected crfs with gaussian edge
  potentials.
\newblock In J.~Shawe-Taylor, R.~S. Zemel, P.~L. Bartlett, F.~Pereira, and
  K.~Q. Weinberger, editors, {\em Advances in Neural Information Processing
  Systems 24}, pages 109--117. Curran Associates, Inc., 2011.

\bibitem{krause2015unreasonable}
J.~Krause, B.~Sapp, A.~Howard, H.~Zhou, A.~Toshev, T.~Duerig, J.~Philbin, and
  F.~Li.
\newblock The unreasonable effectiveness of noisy data for fine-grained
  recognition.
\newblock {\em CoRR}, abs/1511.06789, 2015.

\bibitem{krizhevsky2012imagenet}
A.~Krizhevsky, I.~Sutskever, and G.~E. Hinton.
\newblock Imagenet classification with deep convolutional neural networks.
\newblock In {\em Advances in neural information processing systems}, pages
  1097--1105, 2012.

\bibitem{lam2017fine}
M.~Lam, B.~Mahasseni, and S.~Todorovic.
\newblock Fine-grained recognition as hsnet search for informative image parts.
\newblock In {\em 2017 {IEEE} Conference on Computer Vision and Pattern
  Recognition, {CVPR} 2017, Honolulu, HI, USA, July 21-26, 2017}, pages
  6497--6506, 2017.

\bibitem{Lam_17_HSNet}
M.~Lam, B.~Mahasseni, and S.~Todorovic.
\newblock Fine-grained recognition as hsnet search for informative image parts.
\newblock In {\em 2017 {IEEE} Conference on Computer Vision and Pattern
  Recognition, {CVPR} 2017, Honolulu, HI, USA, July 21-26, 2017}, pages
  6497--6506, 2017.

\bibitem{Li2018ExplicitIB}
X.~Li, Y.~Grandvalet, and F.~Davoine.
\newblock Explicit inductive bias for transfer learning with convolutional
  networks.
\newblock In {\em ICML}, 2018.

\bibitem{lin2017feature}
T.-Y. Lin, P.~Doll{\'a}r, R.~B. Girshick, K.~He, B.~Hariharan, and S.~J.
  Belongie.
\newblock Feature pyramid networks for object detection.
\newblock In {\em CVPR}, volume~1, page~4, 2017.

\bibitem{Liu16FCAN}
X.~Liu, T.~Xia, J.~Wang, and Y.~Lin.
\newblock Fully convolutional attention localization networks: Efficient
  attention localization for fine-grained recognition.
\newblock {\em CoRR}, abs/1603.06765, 2016.

\bibitem{Lu_2018_CVPR}
C.~Lu, H.~Su, Y.~Li, Y.~Lu, L.~Yi, C.-K. Tang, and L.~J. Guibas.
\newblock Beyond holistic object recognition: Enriching image understanding
  with part states.
\newblock In {\em The IEEE Conference on Computer Vision and Pattern
  Recognition (CVPR)}, June 2018.

\bibitem{ren2015faster}
S.~Ren, K.~He, R.~Girshick, and J.~Sun.
\newblock Faster r-cnn: Towards real-time object detection with region proposal
  networks.
\newblock In {\em Advances in neural information processing systems}, pages
  91--99, 2015.

\bibitem{Ristani_2018_CVPR}
E.~Ristani and C.~Tomasi.
\newblock Features for multi-target multi-camera tracking and
  re-identification.
\newblock In {\em The IEEE Conference on Computer Vision and Pattern
  Recognition (CVPR)}, June 2018.

\bibitem{russakovsky2015imagenet}
O.~Russakovsky, J.~Deng, H.~Su, J.~Krause, S.~Satheesh, S.~Ma, Z.~Huang,
  A.~Karpathy, A.~Khosla, M.~Bernstein, et~al.
\newblock Imagenet large scale visual recognition challenge.
\newblock {\em International Journal of Computer Vision}, 115(3):211--252,
  2015.

\bibitem{simon2015neural}
M.~Simon and E.~Rodner.
\newblock Neural activation constellations: Unsupervised part model discovery
  with convolutional networks.
\newblock In {\em Proceedings of the IEEE International Conference on Computer
  Vision}, pages 1143--1151, 2015.

\bibitem{vgg}
K.~Simonyan and A.~Zisserman.
\newblock Very deep convolutional networks for large-scale image recognition.
\newblock {\em CoRR}, abs/1409.1556, 2014.

\bibitem{simonyan2014very}
K.~Simonyan and A.~Zisserman.
\newblock Very deep convolutional networks for large-scale image recognition.
\newblock In {\em International Conference on Learning Representations}, 2015.

\bibitem{Sun_2018_CVPR}
C.~Sun, D.~Wang, H.~Lu, and M.-H. Yang.
\newblock Learning spatial-aware regressions for visual tracking.
\newblock In {\em The IEEE Conference on Computer Vision and Pattern
  Recognition (CVPR)}, June 2018.

\bibitem{sun18}
M.~Sun, Y.~Yuan, F.~Zhou, and E.~Ding.
\newblock Multi-attention multi-class constraint for fine-grained image
  recognition.
\newblock In {\em Computer Vision - {ECCV} 2018 - 15th European Conference,
  Munich, Germany, September 8-14, 2018, Proceedings, Part {XVI}}, pages
  834--850, 2018.

\bibitem{sutton2012introduction}
C.~Sutton, A.~McCallum, et~al.
\newblock An introduction to conditional random fields.
\newblock {\em Foundations and Trends{\textregistered} in Machine Learning},
  4(4):267--373, 2012.

\bibitem{szegedy2015going}
C.~Szegedy, W.~Liu, Y.~Jia, P.~Sermanet, S.~Reed, D.~Anguelov, D.~Erhan,
  V.~Vanhoucke, and A.~Rabinovich.
\newblock Going deeper with convolutions.
\newblock In {\em Proceedings of the IEEE conference on computer vision and
  pattern recognition}, pages 1--9, 2015.

\bibitem{Szegedy2016googlenet}
C.~Szegedy, V.~Vanhoucke, S.~Ioffe, J.~Shlens, and Z.~Wojna.
\newblock Rethinking the inception architecture for computer vision.
\newblock In {\em Proceedings of IEEE Conference on Computer Vision and Pattern
  Recognition,}, 2016.

\bibitem{Teng_2017_ICCV}
Z.~Teng, J.~Xing, Q.~Wang, C.~Lang, S.~Feng, and Y.~Jin.
\newblock Robust object tracking based on temporal and spatial deep networks.
\newblock In {\em The IEEE International Conference on Computer Vision (ICCV)},
  Oct 2017.

\bibitem{van17}
G.~Van~Horn, O.~Mac~Aodha, Y.~Song, Y.~Cui, C.~Sun, A.~Shepard, H.~Adam,
  P.~Perona, and S.~Belongie.
\newblock The inaturalist species classification and detection dataset.
\newblock In {\em Proceedings of the IEEE Conference on Computer Vision and
  Pattern Recognition}, Salt Lake City, UT, 2018.

\bibitem{Wang_2018_CVPR}
X.~Wang, S.~You, X.~Li, and H.~Ma.
\newblock Weakly-supervised semantic segmentation by iteratively mining common
  object features.
\newblock In {\em The IEEE Conference on Computer Vision and Pattern
  Recognition (CVPR)}, June 2018.

\bibitem{wang16}
Y.~Wang, V.~I. Morariu, and L.~S. Davis.
\newblock Learning a discriminative filter bank within a cnn for fine-grained
  recognition.
\newblock In {\em The IEEE Conference on Computer Vision and Pattern
  Recognition (CVPR)}, June 2018.

\bibitem{wang2017multi}
Z.~Wang, T.~Chen, G.~Li, R.~Xu, and L.~Lin.
\newblock Multi-label image recognition by recurrently discovering attentional
  regions.
\newblock In {\em Proceedings of the IEEE Conference on Computer Vision and
  Pattern Recognition}, pages 464--472, 2017.

\bibitem{WelinderEtal2010}
P.~Welinder, S.~Branson, T.~Mita, C.~Wah, F.~Schroff, S.~Belongie, and
  P.~Perona.
\newblock {Caltech-UCSD Birds 200}.
\newblock Technical Report CNS-TR-2010-001, California Institute of Technology,
  2010.

\bibitem{Yu_2018_ECCV}
C.~Yu, X.~Zhao, Q.~Zheng, P.~Zhang, and X.~You.
\newblock Hierarchical bilinear pooling for fine-grained visual recognition.
\newblock In {\em The European Conference on Computer Vision (ECCV)}, September
  2018.

\bibitem{zfnet}
M.~D. Zeiler and R.~Fergus.
\newblock Visualizing and understanding convolutional networks.
\newblock In {\em Computer Vision - {ECCV} 2014 - 13th European Conference,
  Zurich, Switzerland, September 6-12, 2014, Proceedings, Part {I}}, pages
  818--833, 2014.

\bibitem{zhang2014part}
N.~Zhang, J.~Donahue, R.~Girshick, and T.~Darrell.
\newblock Part-based r-cnns for fine-grained category detection.
\newblock In {\em European conference on computer vision}, pages 834--849.
  Springer, 2014.

\bibitem{zhang2013deformable}
N.~Zhang, R.~Farrell, F.~Iandola, and T.~Darrell.
\newblock Deformable part descriptors for fine-grained recognition and
  attribute prediction.
\newblock In {\em Proceedings of the IEEE International Conference on Computer
  Vision}, pages 729--736, 2013.

\bibitem{zheng2017learning}
H.~Zheng, J.~Fu, T.~Mei, and J.~Luo.
\newblock Learning multi-attention convolutional neural network for
  fine-grained image recognition.
\newblock In {\em {IEEE} International Conference on Computer Vision, {ICCV}
  2017, Venice, Italy, October 22-29, 2017}, pages 5219--5227, 2017.

\bibitem{zheng17}
H.~Zheng, J.~Fu, T.~Mei, and J.~Luo.
\newblock Learning multi-attention convolutional neural network for
  fine-grained image recognition.
\newblock In {\em {IEEE} International Conference on Computer Vision, {ICCV}
  2017, Venice, Italy, October 22-29, 2017}, pages 5219--5227, 2017.

\bibitem{zhou2016learning}
B.~Zhou, A.~Khosla, A.~Lapedriza, A.~Oliva, and A.~Torralba.
\newblock Learning deep features for discriminative localization.
\newblock In {\em Proceedings of the IEEE Conference on Computer Vision and
  Pattern Recognition}, pages 2921--2929, 2016.

\end{thebibliography}
}

\end{document}